\documentclass[letterpaper]{article} 
\usepackage{aaai25}  
\usepackage{times}  
\usepackage{helvet}  
\usepackage{courier}  
\usepackage{float}
\usepackage{graphicx}
\usepackage{bm}
\usepackage{color}
\usepackage{amsmath}
\usepackage{multirow}

\usepackage{amsmath,amsfonts,amssymb}

\usepackage{graphicx}
\usepackage[hyphens]{url}  
\usepackage{graphicx} 
\usepackage{natbib}  
\usepackage{caption} 
\usepackage{algorithm}
\usepackage{algorithmic}

\usepackage{newfloat}
\usepackage{listings}
\DeclareCaptionStyle{ruled}{labelfont=normalfont,labelsep=colon,strut=off} 
\floatstyle{ruled}
\newfloat{listing}{tb}{lst}{}
\floatname{listing}{Listing}
\usepackage[utf8]{inputenc} 
\usepackage[T1]{fontenc}    

\title{Distance-Forward Learning: Enhancing the Forward-Forward Algorithm Towards High-Performance On-Chip Learning}
\author{
    Yujie Wu$^{1,*}$, Siyuan Xu$^{2,3,*}$, Jibin Wu$^{1, 4}$, Lei Deng$^{5}$, Mingkun Xu$^{5}$, Qinghao Wen$^{3}$, Guoqi Li$^{2,3,\dagger}$\\
}
\affiliations{
    \textsuperscript{\rm 1}Department of Computing, The Hong Kong Polytechnic University, Hong Kong, China\\
        \textsuperscript{\rm 2}Institute of Automation, Key Laboratory of Brain Cognition and Brain-inspired Intelligence Technology, Chinese Academy
of Sciences, Beijing, China\\
            \textsuperscript{\rm 3}School of Artificial Intelligence, University of Chinese Academy of Sciences, Beijing, China\\
        \textsuperscript{\rm 4}Department of Data Science and Artificial Intelligence, The Hong Kong Polytechnic University, Hong Kong, China\\
        \textsuperscript{\rm 5}Center for Brain Inspired Computing Research, Department of Precision Instrument, Tsinghua University, Beijing, China\\
         \textsuperscript{\rm $\dagger$}guoqi.li@ia.ac.cn\\
}

\usepackage{bibentry}

\begin{document}

\maketitle

\begin{abstract}
The Forward-Forward (FF) algorithm was recently proposed as a local learning method to address the limitations of backpropagation (BP),  
offering biological plausibility along with memory-efficient and highly parallelized computational benefits.
However, it suffers from suboptimal performance and poor generalization, largely due to inadequate theoretical support and a lack of effective learning strategies.  
In this work, we reformulate FF using distance metric learning and propose a distance-forward algorithm (DF)  to improve FF performance in supervised vision tasks while preserving its local computational properties, making it competitive for efficient on-chip learning. To achieve this, we reinterpret FF through the lens of centroid-based metric learning and develop a goodness-based N-pair margin loss to facilitate the learning of discriminative features. Furthermore,  we integrate layer-collaboration local update strategies to reduce information loss caused by greedy local parameter updates. Our method surpasses existing FF models and other advanced local learning approaches, with accuracies of 99.7\% on MNIST, 88.2\% on CIFAR-10, 59\% on CIFAR-100, 95.9\% on SVHN, and 82.5\% on ImageNette, respectively. Moreover, it achieves comparable performance with less than 40\% memory cost compared to BP training, while exhibiting stronger robustness to multiple types of hardware-related noise, demonstrating its potential for online learning and energy-efficient computation on neuromorphic chips.
\end{abstract}
\renewcommand{\thefootnote}{\fnsymbol{footnote}}
\footnotetext[1]{These authors contributed equally to this work.}

\section{Introduction}
Most current deep learning algorithms are trained using backpropagation (BP) in an end-to-end manner \cite{ sohn2016improved,jaiswal2020survey,rippel2015metric,qi2017contrastive,li2021contrastive}, where training losses are computed at the top layer, and weight updates are derived based on gradients flowing downwards. This process introduces notorious update lock issues and suffers from two critical drawbacks for practical on-chip computations. Firstly, it incurs high memory costs due to the need to store intermediate activations of every layer for the computation of gradients. Secondly, it slows the training speed as each layer depends on the gradient calculations from preceding layers. This computational feature also limits parallel distributed processing capabilities and burdens inner-core communications in many-core hardware architectures, such as emerging neuromorphic chips, preventing highly efficient implementations.

On the other hand, the human brain performs synaptic learning in a more efficient, localised manner without waiting for neurons in other brain regions to complete their processes. Recognizing this efficient alternative, Hinton proposed the forward-forward (FF) algorithm \cite{hinton2022forward}, which provides an effective layer-wise learning method that replaces traditional backpropagation with two forward passes. Similar to biological neural systems, the learning process of FF is based on directly adjusting neuron activities—either enhancing or reducing activity—in response to different types of incoming patterns. Crucially, FF does not require perfect knowledge of the forward pass computations, allowing learning to proceed even if some network modules are unknown. Moreover, from a hardware implementation perspective, FF eliminates the necessity to store intermediate activations after each module's computation which significantly reduces memory requirements during training. This facilitates model parallelism in many deep network architectures for faster training and inference.

Despite its considerable computational potential, the FF  still  struggles with poor generalization across many complex datasets. Several approaches have been proposed to enhance FF from different perspectives, such as employing group convolutional operations \cite{papachristodoulou2024convolutional}, integrating learnable embedding representations for label information \cite{dong2018triplet}, adapting to edge applications\cite{pau2023suitability, baghersalimi2023layer},  or applying contrastive learning techniques \cite{aghagolzadeh2024marginal,ahamed2023forward}.
However, none have fully exploited the theoretical interpretation of FFs, and their performances still fail to compete with other advanced local learning methods \cite{wang2021revisiting,journe2022hebbian,ma2023scaling}. More importantly, a comprehensive evaluation of the practical computational advantages of FF-based methods is still lacking.

In this paper, we propose a distance-forward (DF) method to enhance FF for high-performance on-chip applications by using distance metric learning methods and integrating layer-collaboration local weight update strategies. The distance metric space framework offers a transparent geometrical interpretation of FF and its variants, helping us understand the computational principles and guiding our model design.  By exploiting layer-collaboration local gradient update strategies, we aim to offer flexibility in balancing task accuracy and computation cost.
The proposed method has been extensively evaluated on six image classification datasets with different scales, using a deeper layer structure beyond that the original FF considered.  Our three key contributions are included:
\begin{itemize}
\item We provide a geometrical analysis to reformulate FFs from the centroid-based metric learning perspective, bridging the gap between the two techniques. By revisiting metric learning, we delineate the similarities and differences between FF and existing methods. This facilitates understanding of the key computational principles and enlightens readers on how to proceed with further model designs.
\item Building upon the analysis,  we propose a DF model by leveraging a goodness-based margin loss and an N-pair data mining technique. Furthermore, we integrate different layer-collaboration local gradient update strategies in DF. The proposed model can not only learn hierarchical distance-based representation across multiple layers but also provide flexible choices in balancing task accuracy and practical computation cost.
\item  We comprehensively evaluate the DF on accuracy, robustness and computational overhead. The DF outperforms  FF-based models and other advanced local learning methods on versatile image classification tasks, demonstrating significantly better performance. Furthermore,
it  achieves comparable performance with less than 40\% memory cost compared to BP training, while exhibiting stronger robustness to multiple hardware-related sources of noise, highlighting its potential suitability for online learning and neuromorphic chips.
 \end{itemize}
 
\section{Background and related work}

\textbf{Preliminary for the forward-forward algorithm.}
The core idea of the FF is to  replace the forward and backward passes of backpropagation with two forward passes and manipulate the optimization of the goodness function for two types of data (i.e., positive and negative data) with opposite objectives. For the constructed positive data,   the FF training encourages adjustment of the weights to increase the goodness in every hidden layer. Conversely, for the negative data,   it adjusts the weights to decrease the goodness function.

For the supervised image classification tasks, FF manipulates the input images to generate positive and negative samples.
For each input image $\bm{x}\in\mathbb{R}^{m\times 1}$, it replaces the first $K$ pixels of  $\bm{x}$ with the correct (positive) or incorrect (negative)  one-hot label $\bm{y}\in\mathbb{R}^{K\times 1}$.  This process creates modified patterns denoted as $\bm{x^*}$, where $* \in \{pos, neg\}$ indicates whether the label vector is positive or negative. The goodness function $g$ can be formalized as follows:

\begin{equation}
\begin{small}
\bm{v}^{*} = \bm{W}\bm{x^*}, \quad 
    g^{*} = \Vert \bm{v}^{*} \Vert^2_2, \quad   Loss = \sigma(g^{pos}-\theta) + \sigma(\theta -g^{neg}),  
\label{Eq: goodness function}
\end{small}
\end{equation}
where $\bm{W}\in\mathbb{R}^{n\times m}$ denotes the matrix connecting the previous layer to the current layers, $\bm{v}$ denotes the weighted input sum, and $\theta$ is a threshold parameter that modulates the sensitivity to goodness. Given the expression of the goodness function, FF employs the negative log-sigmoid function $\sigma(x)=\log(1 + \exp(-x))$, optimizing the network parameters through simultaneously adjusting $\sigma(g^{pos})$ and  $\sigma(g^{neg})$ in opposite directions.

\textbf{Contrastive loss and distance metric learning.}
Distance metric learning (or simply, metric learning) constructs task-specific distance space so that data samples from the same class are close together in the metric space, while keeping data from different classes far apart. 
Designing a suitable contrastive loss (CL) is fundamental for metric learning. Building on this framework,
triplet loss \cite{dong2018triplet} enhances the model's sensitivity by evaluating the relative distances between an anchor pattern and both a positive and a negative pattern pair. N-pair loss \cite{sohn2016improved,jaiswal2020survey} further broadens the approach by simultaneously contrasting multiple negative samples against a single positive sample. In addition to evaluating the relative distances between pairs of patterns, another relevant research direction adjusts the representation distance between the input patterns and specific centres  \cite{rippel2015metric, qi2017contrastive, li2021contrastive}, which guides the distance representation of different types of patterns. Despite their effectiveness, the majority of CL approaches are built upon end-to-end learning and transmit the representation of the input pattern across layers rather than the distance features, which we will delineate in comparison to FF later.

\textbf{Block-wised local learning methods.}
Another relevant research direction applies CL methods to locally supervised learning tasks, such as allowing gradients to be backpropagated within specific network blocks~\cite{wang2021revisiting,aghagolzadeh2024marginal}, and using iterative gradient update strategies for different block modules  \cite{xiong2020loco,ma2023scaling}. However, these works rely on auxiliary classifiers for decoding, which typically involve multi-layer structures, heavily increasing computational and memory costs. One recent work \cite{ahamed2023forward} applied triplet loss to block-wise learning and relate it with FF. However, it ignores the goodness-based fundamental features of FF, and its performance has yet to reach the state of the art.

\textbf{Bio-inspired local learning methods.}
  Various local training methods have been developed as alternatives to BP. One approach is modifying BP's feedback circuits. Feedback alignment  \cite{lillicrap2016random} provides a more biologically plausible alternative to BP by replacing each of BP's feedback connections with a random matrix. Direct Feedback Alignment (DA) \cite{nokland2016direct} further simplifies weight updates by employing a random direct feedback matrix and avoiding non-local computations. Unlike bio-plausible BP variants, competitive Hebbian learning allows neurons to compete for activation, favoring those most responsive to input \cite{miconi2021hebbian}. Recent progress in soft Hebbian learning \cite{journe2022hebbian} and predictive Hebbian learning \cite{halvagal2023combination} further unveils the potential of bio-inspired learning rules to adapt to deeper layers. However, a noticeable performance gap still exists when applying these techniques to complex AI tasks, particularly when compared to end-to-end BP training. 
  
  Equilibrium propagation (EP) methods\cite{zenke2022hEP,Scellier2024CEP} provide another promising local learning method, which updates the network parameters by employing a dynamic process where the network iterates towards an equilibrium state when presented with perturbation. Despite its bio-plausibility, EP methods suffer from slow convergence and require additional computational resources to reach precise equilibrium states, making them less efficient for on-chip computing.

 \begin{figure*}[!htp]
    \centering
    \includegraphics[width =0.9\textwidth]{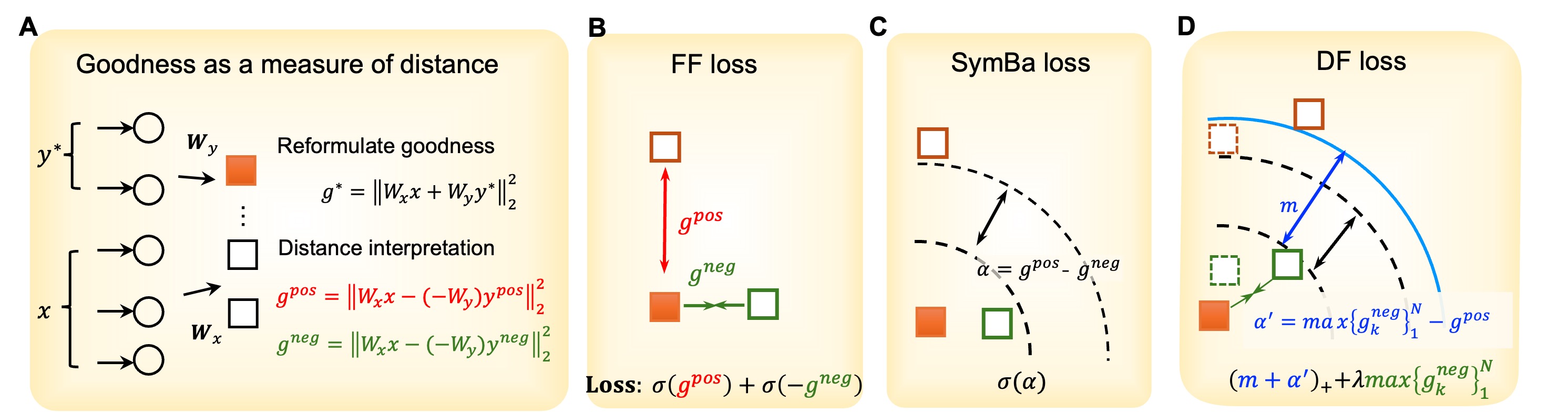}
    \caption{\textbf{Illustration of different forward-forward-based methods in a distance metric space.}\small (A) The goodness function, $g^{pos/neg}$, as proposed by \cite{hinton2022forward}, can be formalized as measuring an $L_2$ distance between $\bm{Wx}$ and $\bm{-W}_y\bm{y}^{pos/neg}$. (B) FF essentially operates on two absolute distances, $g^{pos}$ and $g^{neg}$. $\theta$ in Eq. \ref{Eq: goodness function} is set to zero and thus simplified here for illustration. Here, $\sigma(x)$ represents a negative log-sigmoid function. Optimizing the goodness function can thereby be interpreted as adjusting the distance between the projections of input patterns and anchor patterns (i.e., label vectors).
    (C) The SymBa  loss is proposed to balance the positive and negative losses. It can be interpreted as manipulating the relative discrepancy between $g^{pos}$ and $g^{neg}$ in the goodness-constructed metric space. (D) The proposed DF combines both relative and absolute distances and mines the distances among several positive and negative samples to facilitate the learning of discriminative features. It uses support data points (see solid boxes)—the most representative data points whose distance exceeds a given margin $m$—to calculate a margin loss. This ensures that positive samples are closer to the anchor pattern than negative samples by $m$. Additionally, it includes regularization to further reduce the absolute distance represented by the maximum goodness function of negative samples (see solid green boxes), effectively ensuring all negative samples are kept far away from any anchors. $\lambda$: the weighted coefficient.}
    \label{fig:fig1-motivation}
\end{figure*}

\section{Proposed approach}
 In this section, we reformulate FF and analyze its similarities and distinctive features through the lens of distance metric learning. Building on this foundation, we first propose two goodness-based CL techniques to enhance  performance.  Subsequently, we further elaborate on the two local update strategies to alleviate the overly greedy issues of local learning while preserving computational parallelism. Finally, we present the overall architecture of our model.

\subsection{Formulating FF from distance metric learning}
\label{section:3.1-FF-reformulation}
 There are two fundamental operations employed in conventional metric learning framework: a network first extracts hierarchical representations through multi-layer network modules (denoted the network mapping by $f_\theta$) and then calculates distances within a specific distance metric space between the representations of the input patterns $f_\theta(\bm{x})$ and a specific anchor pattern $f_\theta(\bm{c})$. The anchor pattern can be constructed using different methods, such as sampling another input pattern (i.e., positive sample \cite{dong2018triplet,jaiswal2020survey}) or constructing a cluster centroid representing the central tendency of samples belonging to the same class \cite{yang2016joint,qi2017contrastive,cai2023center} (referred to as \textit{centroid-based metric learning} below). Typically, the two operations can be formulated as
 \begin{equation}
     D_{CL} = F(d(f_\theta(\bm{x}),f_\theta(\bm{c}))).
 \end{equation}
Here $F$ denotes a general format of contrastive loss (e.g., triplet loss) manipulating on the discrepancy $d(f_\theta(\bm{x}),f_\theta(\bm{c}))$.
By analyzing the hidden activations in the first layer of the FF (see Figure \ref{fig:fig1-motivation}), we demonstrate in the following that applying FF to a two-layer network aligns with typical centroid-based metric learning methods \cite{rippel2015metric, qi2017contrastive, li2021contrastive}, whereas extending FF to multi-layer architectures introduces a specific prototype of distance-forward metric learning that initially creates a distance representation in the shallow layers and refines this representation in deeper layers.

\textbf{Relating FF with centroid-based metric learning in a two-layer structure.} 
Our idea stems from a reconstruction of the weight matrix \( \bm{W} \in \mathbb{R}^{n \times m} \) , which divides $\bm{W}$ into two parts: \( \bm{W}_x \in \mathbb{R}^{n \times m} \) for the input patterns \( \bm{x} \) and \( \bm{W}_y \in \mathbb{R}^{n \times K} \) for the labels \( \bm{y}^* \).  Here we set the first $K$ rows of  $\bm{W}_x$ to zeros to cancel out the impact of the first K pixels on $\bm{x}^*$ employed in Eq. \ref{Eq: goodness function}. Then, the weighted sum $\bm{v}$ can  be formulated as:

\begin{equation}
\begin{small}
\bm{v}^{*} = \bm{W}_x\bm{x}+\bm{W}_y\bm{y}^{*}, \quad *\in \{pos, neg\}.
\label{Eq: goodness v}
\end{small}
\end{equation}
 Reformulating Eq. \ref{Eq: goodness v} by introducing $\bm{W^-}_y:=-\bm{W}_y$ gives a clearer distance-based expression:
 
\begin{equation}
g^{pos} = \Vert \bm{W}_x\bm{x}-\bm{W}^-_y\bm{y}^{pos} \Vert^2_2, \quad g^{neg} = \Vert \bm{W}_x\bm{x}-\bm{W^-}_y \bm{y}^{neg} \Vert^2_2.
\label{Eq: goodness function-L2}
\end{equation}
The loss function of Eq. \ref{Eq: goodness function} then optimizes the two distances in opposite directions. This formulation indicates that the original goodness function calculates the distance between the projection of input patterns $\bm{W}_x \bm{x}$ and different labels $\bm{W^-}_y \bm{y}^{neg}$ or $\bm{W^-}_y \bm{y}^{pos}$. Thus, optimizing the goodness function essentially adjusts these absolute Euclidean distances. 
Observing that the projection of label information remains consistent across different patterns, it acts as an anchor vector to guide the metric learning process.  This approach aligns with the  principles of previous centroid-based metric learning methods, where distances between input samples and a constructed centroid (e.g., K-mean-based clustering \cite{yang2016joint}) are used for adjusting representation distances. 
 
This reformulation offers a fresh perspective on FF and its variants, and the following items elucidate its implications:

\begin{itemize}
    \item \textbf{Geometrical interpretation of FF and its variants.} FF can be interpreted as manipulating the absolute distance in an $L_2$ metric space.  Follow-up studies \cite{lee2023symba, dooms2023trifecta} further improve performance and convergence of FF by using a new loss function $\sigma(g^{pos}-g^{neg})$. As indicated by Figure \ref{fig:fig1-motivation}C, this function can be interpreted as optimizing the relative distance between $g^{pos}$ and $g^{neg}$, which avoids being overly influenced by the scale of activation and balances the distances between positive and negative samples more effectively.
    \item \textbf{Effectiveness of minimizing/maximizing goodness.}  As observed by Hinton \cite{hinton2022forward}, optimizing the goodness function of positive samples in opposite directions leads to comparable classification accuracy. Figure \ref{fig:fig1-motivation}B provides a clear explanation: both strategies, pushing the positive samples closer to or further away from the anchor vectors (i.e., the label vector), are effective for learning the discriminative distance distribution of different samples.
    \item \textbf{Effectiveness of label encoding.} Studies  \cite{kohan2023signal,dooms2023trifecta}  suggest that learnable embedding representations $\bm{y}$ enhances goodness discrepancies and leads to better performance. This feature can be supported from the perspective of metric learning, which indicates that well-separated representations for centroid patterns benefit the learning of discriminative representations.
    \end{itemize}
 
The above analysis illustrates the similarity between FFs and centroid-based metric learning. In the following section, we will provide an analysis of the unique features of FF when applied to multi-layer architecture.

\textbf{Refining distance-based representation through multiple-layer structure.} Compared with conventional CL techniques, FF inverts the order of the above two operations: it begins by deriving distance-based representations in the first hidden layer, as illustrated previously, and subsequently propagates these distance-based representations to deeper layers. This process can be formalized by the following equation:

\begin{equation}
     D_{FF} = F(f_\theta(d(\bm{x},\bm{c}))).
 \end{equation}

Considering that the primary objective of supervised contrastive learning is to extract discriminative features from input data to enhance pattern recognition, we hypothesize that the multi-layer architecture of FF can gradually refine the distance metric initially established in the shallow layers through multi-layer computation and thereby attain more discriminative features. This hypothesis will be empirically tested in the experimental section to verify the efficacy of multi-layer enhancement.

\subsection{Goodness-based margin loss for enhanced distance representation}
\label{section:3.2-FF-loss}
We have interpreted goodness as a distance measure and conceptualized the FF within a distance metric space. From this perspective, the activation of the first hidden layer represents the discrepancy between projections of  input patterns and labels, and the information propagation across layers can further refine this discrepancy representation for enhancing feature discriminability. Inspired by the geometrical interpretation and principles from the triplet loss \cite{dong2018triplet}, we propose a goodness-based margin loss with an additional regularization term on the goodness function:

\begin{equation}
    L = \max(m^++g^{neg}-g^{pos},0)+\lambda g^{neg},
    \label{Eq:margin-loss}
\end{equation}
where $\lambda$ denotes the coefficient of the regularization term. 
\begin{figure*}[!htp]
    \centering
    \includegraphics[width =0.75\textwidth]{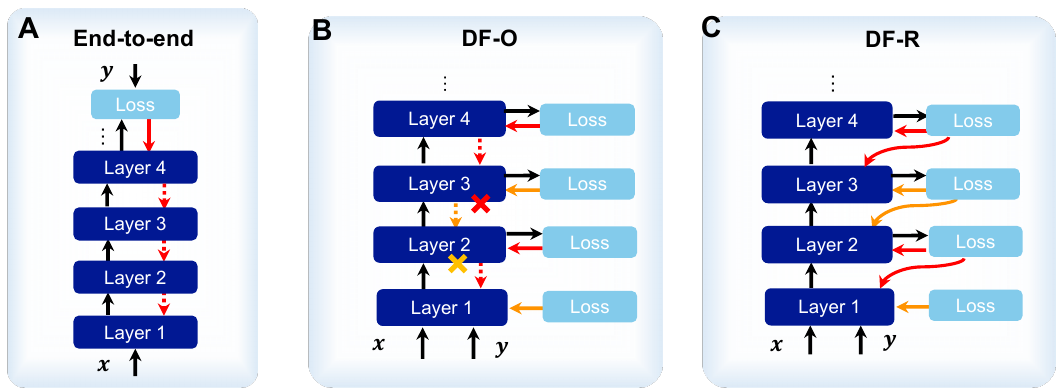}
    \caption{Scheme of different gradient update strategies. Unlike end-to-end approaches, DF-O calculates the loss for every layer and blocks information backpropagation across multiple layers (see  red dashed line). DF-R further integrates random feedback to replace the indirect backward circuit, achieving higher computational parallelism and better bio-plausibility. }
    \label{fig:fig2-local-update-illustration}
\end{figure*}

Instead of penalizing the discrepancy for every positive and negative pair, the margin loss seeks to impose stronger penalties on those support vectors (see solid boxes in Figure \ref{fig:fig1-motivation}D) that can effectively separate the positive and negative goodness functions.  From the perspective of metric space, this encourages positive samples to be closer to the anchor of the label projection than negative samples  by a specified margin $m^+$. Furthermore, the regularization further reduces the absolute distance denoted by the goodness function of negative samples, ensuring that negative samples are kept far away from any projections of labels. 

\subsection{N-pair negative sample mining for label balance}
\label{section:3.3-N-pair}
 The FF and its previous variants learn from  single pairs of positive and negative samples. For a given input query sample, these methods compare it with one negative sample, neglecting negative samples from  other classes.  This strategy can lead to the label-imbalance issue in complex tasks involving many categories because individual updates may bias the distance between a single pair of positive and negative samples, making convergence unstable and slow. Hence, employing multiple N-negative samples for learning could be beneficial for alleviating the convergence problem.   Notably,  the concept of employing an N-pair loss function is not new; it has been  proposed in metric learning for many end-to-end BP methods \cite{sohn2016improved,jaiswal2020survey}. Here we adapt this idea to the local learning context, aiming to enhance the model's feature extraction capabilities for complex tasks.
 
To this end, we employ multiple negative samples for calculating the loss function of the goodness function.  By leveraging the multiple negative samples, we  derive a more comprehensive loss function built on Eq. \ref{Eq:margin-loss}:

\begin{equation}
\begin{small}
    L = \max(m^++max\{g_k^{neg}\}_{k=1}^{N}-g^{pos}, 0)+\lambda max(\{g_k^{neg}\}_{k=1}^{N}), 
    \label{Eq:group-loss}
    \end{small}
\end{equation}
where the lower index $k$ refers to the index of negative samples and $N$ refers to the number of negative samples generated for one positive sample.

\subsection{Local layer collaboration strategies for DF models} 
\label{section:3.4-gradient-strategy}
A large body of local learning methods, including FF, updates weights in a greedy, layerwise manner. This approach can result in the loss of useful features for deeper layers since the update in lower layers is unaware of the higher-level representation needs.  To facilitate effective information communication across layers while preserving the computational superiority of FF,  we develop two weight update strategies, named DF-O and DF-R, based on the methods introduced in \cite{nokland2016direct, xiong2020loco}. These strategies can provide greater flexibility in balancing task accuracy, hardware deployability, and computational cost.

\textbf{DF-O}: First, to achieve high task accuracy, the DF-O employs an overlapping gradient update~(OG) strategy \cite{xiong2020loco} by default. As shown in Figure \ref{fig:fig2-local-update-illustration}B, this strategy iteratively groups two adjacent layers and trains each block locally, allowing gradient information to propagate only across two grouped layers. This method offers a significant advantage over the greedy local update by enabling information flow from the top layer to the bottom layer with only a minimal increase in computational load. A similar method has been successfully used in \cite{dooms2023trifecta} to enhance classification accuracy.

\textbf{DF-R}: Second, to further enhance biological plausibility and facilitate on-chip  implementation, DF-R integrates random direct feedback (FA) connections \cite{nokland2016direct} and the OG strategy. As illustrated in Figure \ref{fig:fig2-local-update-illustration}C, a random feedback connection replaces the back-propagation circuits within the OG update (see dashed red line in Figure \ref{fig:fig2-local-update-illustration}B). It thereby allows the loss to be directly used to update all relevant weights. This approach offers three main benefits beyond task accuracy:  (1) \textit{Higher on-chip computational parallelism}. The fully local update can better leverage the pipeline processing mechanisms of many-core chips and  minimize idle times for processing units. (2) \textit{Reduced volume of data transfer between computation cores}. By limiting the need for extensive communication between cores, DF-R effectively alleviates the  inter-core communication workload. (3) \textit{Increased biological plausibility}.The weight-asymmetric update method breaks the local update-locking required by OG methods and enables real-time weight updates, which align more closely with biological plasticity circuits. Implementation details for the two strategies are provided in Appendix  A.4.

\subsection{Overall architecture of DF models}
 \label{section:3.5-overall architecture}
 To provide a comprehensive understanding, we outline the general training procedures of our methods below.

\textbf{Generation of positive and negative samples.}  We adopt a learnable linear projection for encoding the label information. For simplicity, we use a single-channel learnable embedding for representing label information, keeping the embedding size equal to a single channel of the input image. Positive and negative samples are then generated by concatenating original images with the embedding representation for correct and incorrect labels, respectively. 

\textbf{Training phase.}   We follow  the basic training phase of \cite{hinton2022forward} to ensure consistency. For each positive sample, we generate N pairs of negative samples by randomly selecting N incorrect labels. Subsequently, the positive and negative samples are fed into the network to obtain corresponding activations. We then calculate the goodness function, optimize the loss function   and update the network parameters. 
Batch normalization, as suggested by \cite{dooms2023trifecta}, is performed before calculating the weighted sum for each layer.

\textbf{Decoding and evaluation.} The correct output labels are chosen following Hinton's goodness-based evaluation methods \cite{hinton2022forward}. Specifically, for each testing sample, we evaluate the goodness of different layers in response to combinations of testing samples and candidate labels. The labels with the highest goodness value are selected as output labels.

\section{Experiments}

\subsection{Experimental setup}
We extensively evaluate model performance on six datasets including MNIST, Fashion MNIST (F-MNIST), SVHN, CIFAR-10, CIFAR-100, and Imagenette. The Imagenette is a subset of ImageNet, which is used to facilitate comparison against the advanced local learning method introduced in~\cite{journe2022hebbian}. To facilitate experimental analysis, we use a ten-layer CNN structure for the results presented in Table \ref{table:ann_comparison}, Table \ref{table:models_comparison}, Figures \ref{fig:fig3-computation} C-E, Figure S1 and S2~(see Appendix B for details).  The default optimizer, Adam, with a learning rate of 1e-3 and a cosine scheduler, has been utilized across all experiments. Details of other hyperparameter settings are available in Appendix A.1.
 
\begin{figure}[!htp]
    \centering
    \includegraphics[width =0.42\textwidth]{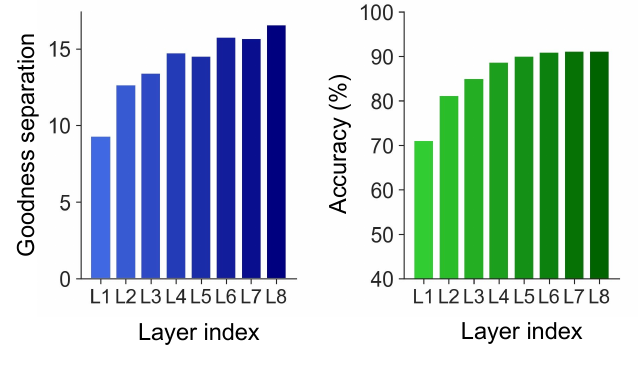}
    \caption{Testing classification accuracy and average separation of goodness on Fashion MNIST using the DF-O. The separation is measured as the difference in goodness between positive and negative training samples. }
    \label{fig:fig2-df-representation}
\end{figure}
\subsection{Comprehensive performance evaluation}
\textbf{Task accuracy.} We evaluate the proposed DF methods against end-to-end BP learning, FF-based variants, and other state-of-the-art non-BP approaches across six datasets.
 As shown in Table \ref{table:ann_comparison}, both DF-R and DF-O models perform well on small-scale datasets and outperform other FF-based models in most evaluation datasets.Furthermore, benefiting from relaxing the constraint and allowing block-wise gradient backpropagation, the DF-O demonstrates comparable performance with other advanced local learning methods, approaching the performance of end-to-end BP learning.  We also notice that CFSE \cite{papachristodoulou2024convolutional} shows a higher accuracy on CIFAR100. By adopting the softmax classifier in their model, CFSE employs a different decoding strategy that may benefit performance while producing a larger number of trainable parameters.

\begin{table*}[htp]
\centering
\small
\setlength{\tabcolsep}{3pt}
\caption{Comparison of test accuracies across various datasets, with mean and standard deviation (std) computed over three independent trials. NR*: No Results reported by previous works on the specific benchmarks. }
\scalebox{0.95}{
\begin{tabular}{ccccccc}
\hline\noalign{\hrule height 1pt}
Model    &   MNIST & F-MNIST & SVHN  & CIFAR10 & CIFAR100 & Imagenette \\ \hline 
BP-CNN   & 99.63 & 93.74        & 95.2  & 90.1    & 60.4     & 85.3       \\ \hline
\multicolumn{7}{c}{Non-backpropagation approaches}                                     \\ \hline
PEPITA \cite{dellaferrera2022error}      &   98.29 & NR            & NR     & 56.33   & 27.56    & NR          \\
DA \cite{nokland2016direct} &   NR     & 91.54            & NR     & 62.7    & 48.03       & NR       \\
DKP \cite{webster2020learning} &   NR     & 91.66   $\pm$ 0.27          & NR     & 64.69 $\pm$ 0.72    & 52.62  $\pm$0.48     & NR       \\
Softhebb \cite{journe2022hebbian} &   NR     & NR            &  NR     & 80.3    & 56       & 81.0       \\
InfoPro \cite{wang2021revisiting}  &   NR     & NR            & 94.86 & 87.25   & NR        & NR          \\ 
\hline
\multicolumn{7}{c}{Forward-Forward-based approaches}                                         \\ \hline
FF-CNN \cite{hinton2022forward}   &   99.4  & NR            &  NR     & 59      & NR        & NR         \\
Symba \cite{lee2023symba}    &   98.48 & NR        & NR & 59.09   & 28.28    & NR          \\
TFF \cite{dooms2023trifecta}    &  99.58 $\pm$ 0.06 & 91.44 $\pm$ 0.49       & 94.31 $\pm$ 0.07 & 83.51 $\pm$ 0.78   & 35.26 $\pm$ 0.23   & NR         \\
CFF \cite{lorberbom2024layer}    &  97.9 & 88.4         & NR & 48.4   & NR   & NR          \\
FFCM \cite{ahamed2023forward}    &  97.12 & 87.64        & NR & 54.48   & NR   & NR          \\
CFSE \cite{papachristodoulou2024convolutional}     &  99.42  & 92.21        & NR     & 78.11    & 51.23    & NR          \\ \hline
\textbf{DF-R} (layer-wise)    & 99.53 &  92.5            & 94.97  & 84.75   & 48.16    &    81.2           \\ 
\textbf{DF-O} (block-wise)   & \textbf{99.70 $\pm$ 0.09} &  \textbf{93.89  $\pm$ 0.25}            & \textbf{95.91 $\pm$ 0.13}  & \textbf{88.15$\pm$0.28}   & \textbf{59.01$\pm$0.35}     &     \textbf{82.5$\pm$0.3}           \\ \hline\noalign{\hrule height 1pt}
\end{tabular}
}
\label{table:ann_comparison}
\end{table*}

\textbf{Hierarchical representations through deep layers.} The foremost question for DF is whether the propagation of distance-based representation can leverage multi-layer architecture to facilitate the learning of discriminative features. To this end, we analyze the representation of goodness across different layers using the DF-O model, as depicted in Figure \ref{fig:fig2-df-representation}. We quantitatively measure the discrepancy in goodness between positive and negative training samples, alongside the testing accuracy derived from the goodness function of each layer. As demonstrated in Figure \ref{fig:fig2-df-representation}, a consistent discrepancy in goodness is observed between the positive and negative samples across tasks in deeper layers. This change in discrepancy correlates with the trend in testing accuracy, providing evidence of the hierarchical representations achieved by  DF.
 
\begin{figure}[!htp]
    \centering
    \includegraphics[width =0.51\textwidth]{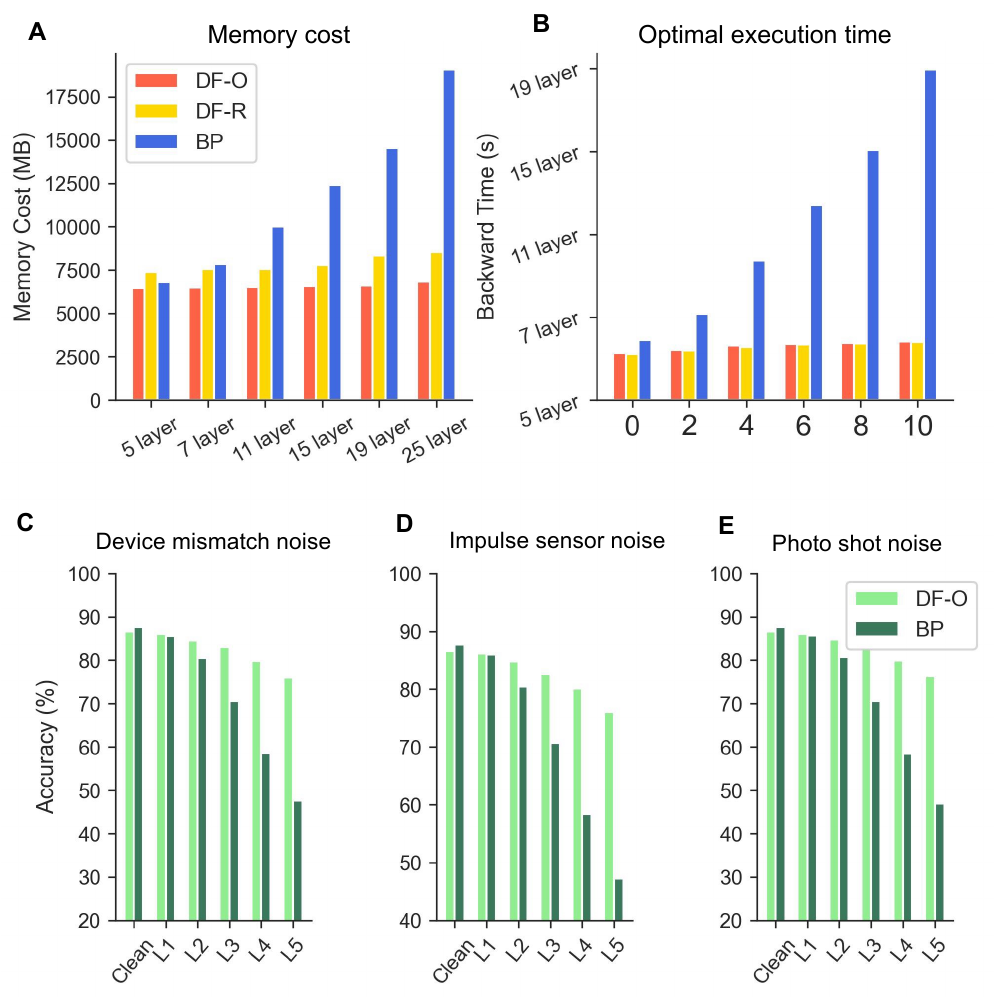}
    \caption{DF inherits the memory-efficient (A) and parallelized computational benefits (B) of FF and exhibits strong robustness to three types of hardware-related noise (C-E) compared with BP methods. L1-L5: specific noise levels.}
    \label{fig:fig3-computation}
\end{figure}

\textbf{Hardware computational efficiency, quantization and robustness to the device-induced noise.} 
The local computation paradigm of DF methods make it hardware friendly. We comprehensively evaluate this feature by examining memory costs, training time, quantization effect, and robustness to hardware-related noise. As illustrated in Figures \ref{fig:fig3-computation}A and B, the end-to-end BP requires storing activations during inference for parameter updates, leading to linear increases in memory costs and training time as the number of network layers grows. Conversely, the memory cost for both DF models remains significantly lower than that for BP, requiring less than about 40\% of the memory cost for an 11-layer architecture and about 60\% for a 25-layer architecture. Additionally, since the parameter update procedure in DF-R is highly parallel, its optimal gradient computation time—estimated by  the backward time in PyTorch (see Appendix A.4)—is lower than that of BP. 
We also note that the two-layer block-wise update strategy does not significantly burden computational parallelism. Nevertheless, it is worth clarifying that for DF-R, if one considers grouping more than two layers for overlapping updates, the computational advantages of DF-R can be more pronounced (see Figure S2).

Unlike ideal simulation environments, device non-ideality issues in analog computing introduce various types of noise, impacting on-chip computations. We evaluate the robustness of  DF methods to diverse hardware-related noises, focusing on three typical noise sources: \textit{Device mismatch noise}, \textit{impulse sensor noise}, and \textit{photon shot noise} \cite{hendrycks2018benchmarking,yang2022training}. \textit{Device mismatch noise}, causing network parameters to deviate from desired values, is simulated by adding Gaussian noise to the gradient of network parameters in each layer during the training phase. \textit{Photon shot noise} inherent in the discrete nature of light, is simulated by introducing Poisson noise into  testing patterns. \textit{Impulse sensor noise}, typical in neuromorphic vision sensors, is simulated by injecting a color analogue of salt-and-pepper noise into testing patterns.  
We compare the DF-O and the vanilla BP models across different types of testing noises  and various noise levels (see Appendix A.3 for implementation details) and measure the testing accuracy. As indicated in Figures \ref{fig:fig3-computation}C-E, while under a clean condition, DF-O performance is slightly worse than that of backpropagation. As the noise levels increase, DF-O achieves consistently higher classification accuracy compared to the BP approach. This superiority is likely due to the crafted margin loss, which facilitates creating  discriminative and robust representations. Additionally, our quantization evaluation shown in  Figure S3 corroborates that our model works well even in 4-bit configuration. Altogether, our validations suggest a stable on-chip performance of the proposed model.

 \begin{table}[!htp]
\centering
\caption{Evaluation of the effectiveness of the N negative pairs using the DF-O on CIFAR10 and CIFAR100. }
\scalebox{0.9}{
\begin{tabular}{c|ccc|c}
\hline\noalign{\hrule height 1pt}
Dataset  & $N$=1 & $N$=3-max &  $N$=9-max & $N$=9-avg \\ \hline
CIFAR10  & 87.01      & 87.71         & 88.15     & 80.3         \\
CIFAR100 & 52.41      & 56.2               & 59.01         & 49.41       \\ \hline\noalign{\hrule height 1pt}
\end{tabular}}
\label{table:models_comparison}
\end{table}

\subsection{Ablation study}
\paragraph{Effectiveness of the N-pair loss design.} We compare the impact of different numbers of negative pairs in calculating the loss function (see Eq. \ref{Eq:group-loss}) on classification accuracy using the DF-O.  Table \ref{table:models_comparison} shows that employing more negative samples consistently improves performance accuracy. Given the larger number of categories in CIFAR100, a notable increase in classification accuracy can be observed on CIFAR100 as the number of N-pairs increases. 
Additionally, we also replace the maximum operation by  using the average of several negative sample projections to calculate the loss in Eq. \ref{Eq:group-loss}. Table \ref{table:models_comparison} suggests that this strategy led to even worse performance, validating our rationality for utilization of N-pair sampling.

\textbf{Impact of regularization coefficients.}  We examine the impact of regularization by varying the regularization coefficients ($\lambda$)  in Figures S1A and B using the DF-O model. Noting that $\lambda=0$ equates to a model without regularization, results  indicate that a suitable $\lambda$ value can achieve higher classification accuracy than the model without regularization, validating the utility of this regularization component.

\textbf{Effectiveness of the goodness-based margin loss.} We compare in Figure S1C the impact of the proposed loss function (Eq. \ref{Eq:group-loss}) against  the original FF on task performance. For fairness, both models are equipped with identical network architectures and OG update strategy.
Results show that the proposed loss function leads to consistent improvements over FF on both Fashion MNIST and CIFAR-10 datasets, substantiating the efficacy of the loss designs.


\section{Conclusions}
\label{section: limitations}
We offer a new perspective of FF from distance metric learning and  present a DF method, which employs a goodness-based N-pair margin loss and integrates layer-collaboration gradient update methods, enabling the extraction of discriminative features while preserving the computation benefits of local learning. Our extensive experiments confirm that the DF models outperform FF-based models, achieving comparable results to other advanced local learning methods.
Furthermore, our quantitative analysis of memory cost, training time, and robustness to multiple hardware-related noise corroborates the potential advantages of the DF in high-performance on-chip computations, making it suitable for energy-efficient cutting-edge applications.
\newpage

\bibliography{reference}

\begin{thebibliography}{32}
\providecommand{\natexlab}[1]{#1}

\bibitem[{Aghagolzadeh and Ezoji(2024)}]{aghagolzadeh2024marginal}
Aghagolzadeh, H.; and Ezoji, M. 2024.
\newblock Marginal Contrastive Loss: A Step Forward for Forward-Forward.
\newblock In \emph{2024 13th Iranian/3rd International Machine Vision and Image Processing Conference}, 1--6. IEEE.

\bibitem[{Ahamed, Chen, and Imran(2023)}]{ahamed2023forward}
Ahamed, M.~A.; Chen, J.; and Imran, A.-A.-Z. 2023.
\newblock Forward-forward contrastive learning.
\newblock \emph{arXiv preprint arXiv:2305.02927}.

\bibitem[{Baghersalimi et~al.(2023)Baghersalimi, Amirshahi, Teijeiro, Aminifar, and Atienza}]{baghersalimi2023layer}
Baghersalimi, S.; Amirshahi, A.; Teijeiro, T.; Aminifar, A.; and Atienza, D. 2023.
\newblock Layer-Wise Learning Framework for Efficient DNN Deployment in Biomedical Wearable Systems.
\newblock In \emph{2023 IEEE 19th International Conference on Body Sensor Networks (BSN)}, 1--4. IEEE.

\bibitem[{Cai, Xiong, and Tian(2023)}]{cai2023center}
Cai, B.; Xiong, P.; and Tian, S. 2023.
\newblock Center Contrastive Loss for Metric Learning.
\newblock \emph{arXiv preprint arXiv:2308.00458}.

\bibitem[{Dellaferrera and Kreiman(2022)}]{dellaferrera2022error}
Dellaferrera, G.; and Kreiman, G. 2022.
\newblock Error-driven input modulation: Solving the credit assignment problem without a backward pass.
\newblock In \emph{International Conference on Machine Learning}, 4937--4955. PMLR.

\bibitem[{Dong and Shen(2018)}]{dong2018triplet}
Dong, X.; and Shen, J. 2018.
\newblock Triplet loss in siamese network for object tracking.
\newblock In \emph{Proceedings of the European conference on computer vision}, 459--474.

\bibitem[{Dooms, Tsang, and Oramas(2023)}]{dooms2023trifecta}
Dooms, T.; Tsang, I.~J.; and Oramas, J. 2023.
\newblock The Trifecta: Three simple techniques for training deeper Forward-Forward networks.
\newblock \emph{arXiv preprint arXiv:2311.18130}.

\bibitem[{Halvagal and Zenke(2023)}]{halvagal2023combination}
Halvagal, M.~S.; and Zenke, F. 2023.
\newblock The combination of Hebbian and predictive plasticity learns invariant object representations in deep sensory networks.
\newblock \emph{Nature Neuroscience}, 26(11): 1906--1915.

\bibitem[{Hendrycks and Dietterich(2018)}]{hendrycks2018benchmarking}
Hendrycks, D.; and Dietterich, T.~G. 2018.
\newblock Benchmarking neural network robustness to common corruptions and surface variations.
\newblock \emph{arXiv preprint arXiv:1807.01697}.

\bibitem[{Hinton(2022)}]{hinton2022forward}
Hinton, G. 2022.
\newblock The forward-forward algorithm: Some preliminary investigations.
\newblock \emph{arXiv preprint arXiv:2212.13345}.

\bibitem[{Jaiswal et~al.(2020)Jaiswal, Babu, Zadeh, Banerjee, and Makedon}]{jaiswal2020survey}
Jaiswal, A.; Babu, A.~R.; Zadeh, M.~Z.; Banerjee, D.; and Makedon, F. 2020.
\newblock A survey on contrastive self-supervised learning.
\newblock \emph{Technologies}, 9(1): 2.

\bibitem[{Journ{\'e} et~al.(2022)Journ{\'e}, Rodriguez, Guo, and Moraitis}]{journe2022hebbian}
Journ{\'e}, A.; Rodriguez, H.~G.; Guo, Q.; and Moraitis, T. 2022.
\newblock Hebbian deep learning without feedback.
\newblock \emph{arXiv preprint arXiv:2209.11883}.

\bibitem[{Kohan, Rietman, and Siegelmann(2023)}]{kohan2023signal}
Kohan, A.; Rietman, E.~A.; and Siegelmann, H.~T. 2023.
\newblock Signal propagation: The framework for learning and inference in a forward pass.
\newblock \emph{IEEE Transactions on Neural Networks and Learning Systems}.

\bibitem[{Laborieux~A(2022)}]{zenke2022hEP}
Laborieux~A, Z.~F. 2022.
\newblock Holomorphic equilibrium propagation computes exact gradients through finite size oscillations.
\newblock \emph{Advances in neural information processing systems}, 35.

\bibitem[{Lee and Song(2023)}]{lee2023symba}
Lee, H.-C.; and Song, J. 2023.
\newblock Symba: Symmetric backpropagation-free contrastive learning with forward-forward algorithm for optimizing convergence.
\newblock \emph{arXiv preprint arXiv:2303.08418}.

\bibitem[{Li et~al.(2021)Li, Hu, Liu, Peng, Zhou, and Peng}]{li2021contrastive}
Li, Y.; Hu, P.; Liu, Z.; Peng, D.; Zhou, J.~T.; and Peng, X. 2021.
\newblock Contrastive clustering.
\newblock In \emph{Proceedings of the AAAI conference on artificial intelligence}, volume~35.

\bibitem[{Lillicrap et~al.(2016)Lillicrap, Cownden, Tweed, and Akerman}]{lillicrap2016random}
Lillicrap, T.~P.; Cownden, D.; Tweed, D.~B.; and Akerman, C.~J. 2016.
\newblock Random synaptic feedback weights support error backpropagation for deep learning.
\newblock \emph{Nature communications}, 7(1): 13276.

\bibitem[{Lorberbom et~al.(2024)Lorberbom, Gat, Adi, Schwing, and Hazan}]{lorberbom2024layer}
Lorberbom, G.; Gat, I.; Adi, Y.; Schwing, A.; and Hazan, T. 2024.
\newblock Layer collaboration in the forward-forward algorithm.
\newblock In \emph{Proceedings of the AAAI Conference on Artificial Intelligence}, volume~38.

\bibitem[{Ma et~al.(2023)Ma, Wu, Si, and Tan}]{ma2023scaling}
Ma, C.; Wu, J.; Si, C.; and Tan, K. 2023.
\newblock Scaling Supervised Local Learning with Augmented Auxiliary Networks.
\newblock In \emph{The Twelfth International Conference on Learning Representations}.

\bibitem[{Miconi(2021)}]{miconi2021hebbian}
Miconi, T. 2021.
\newblock Hebbian learning with gradients: Hebbian convolutional neural networks with modern deep learning frameworks.
\newblock \emph{arXiv preprint arXiv:2107.01729}.

\bibitem[{N{\o}kland(2016)}]{nokland2016direct}
N{\o}kland, A. 2016.
\newblock Direct feedback alignment provides learning in deep neural networks.
\newblock \emph{Advances in neural information processing systems}, 29.

\bibitem[{Papachristodoulou et~al.(2024)Papachristodoulou, Kyrkou, Timotheou, and Theocharides}]{papachristodoulou2024convolutional}
Papachristodoulou, A.; Kyrkou, C.; Timotheou, S.; and Theocharides, T. 2024.
\newblock Convolutional Channel-Wise Competitive Learning for the Forward-Forward Algorithm.
\newblock In \emph{Proceedings of the AAAI Conference on Artificial Intelligence}, volume~38.

\bibitem[{Pau and Aymone(2023)}]{pau2023suitability}
Pau, D.~P.; and Aymone, F.~M. 2023.
\newblock Suitability of forward-forward and pepita learning to mlcommons-tiny benchmarks.
\newblock In \emph{2023 IEEE International Conference on Omni-layer Intelligent Systems (COINS)}, 1--6. IEEE.

\bibitem[{Qi and Su(2017)}]{qi2017contrastive}
Qi, C.; and Su, F. 2017.
\newblock Contrastive-center loss for deep neural networks.
\newblock In \emph{2017 IEEE international conference on image processing}, 2851--2855. IEEE.

\bibitem[{Rippel et~al.(2015)Rippel, Paluri, Dollar, and Bourdev}]{rippel2015metric}
Rippel, O.; Paluri, M.; Dollar, P.; and Bourdev, L. 2015.
\newblock Metric learning with adaptive density discrimination.
\newblock \emph{arXiv preprint arXiv:1511.05939}.

\bibitem[{Scellier et~al.(2024)Scellier, Ernoult, Kendall, and Kumar}]{Scellier2024CEP}
Scellier, B.; Ernoult, M.; Kendall, J.; and Kumar, S. 2024.
\newblock Energy-based learning algorithms for analog computing: a comparative study.
\newblock \emph{Advances in Neural Information Processing Systems}, 36.

\bibitem[{Sohn(2016)}]{sohn2016improved}
Sohn, K. 2016.
\newblock Improved deep metric learning with multi-class n-pair loss objective.
\newblock \emph{Advances in neural information processing systems}, 29.

\bibitem[{Wang et~al.(2020)Wang, Ni, Song, Yang, and Huang}]{wang2021revisiting}
Wang, Y.; Ni, Z.; Song, S.; Yang, L.; and Huang, G. 2020.
\newblock Revisiting Locally Supervised Learning: an Alternative to End-to-end Training.
\newblock In \emph{International Conference on Learning Representations}.

\bibitem[{Webster, Choi et~al.(2020)}]{webster2020learning}
Webster, M.~B.; Choi, J.; et~al. 2020.
\newblock Learning the connections in direct feedback alignment.
\newblock \emph{openreview}.

\bibitem[{Xiong, Ren, and Urtasun(2020)}]{xiong2020loco}
Xiong, Y.; Ren, M.; and Urtasun, R. 2020.
\newblock Loco: Local contrastive representation learning.
\newblock \emph{Advances in neural information processing systems}, 33.

\bibitem[{Yang, Parikh, and Batra(2016)}]{yang2016joint}
Yang, J.; Parikh, D.; and Batra, D. 2016.
\newblock Joint unsupervised learning of deep representations and image clusters.
\newblock In \emph{Proceedings of the IEEE conference on computer vision and pattern recognition}, 5147--5156.

\bibitem[{Yang et~al.(2022)Yang, Wu, Zhang, Chua, Wang, and Li}]{yang2022training}
Yang, Q.; Wu, J.; Zhang, M.; Chua, Y.; Wang, X.; and Li, H. 2022.
\newblock Training spiking neural networks with local tandem learning.
\newblock \emph{Advances in Neural Information Processing Systems}, 35.

\end{thebibliography}

\end{document}